\documentclass{article} 
\usepackage{iclr2027_conference,times}
\iclrfinalcopy

\usepackage{amsmath,amsfonts,bm}

\def\eqref#1{equation~\ref{#1}}

\def\1{\bm{1}}

\DeclareMathAlphabet{\mathsfit}{\encodingdefault}{\sfdefault}{m}{sl}
\SetMathAlphabet{\mathsfit}{bold}{\encodingdefault}{\sfdefault}{bx}{n}

\usepackage{url}
\usepackage{microtype}
\usepackage{graphicx}
\usepackage{wrapfig}
\usepackage{subcaption}
\usepackage{booktabs} 
\usepackage{multirow}
\usepackage{diagbox}
\usepackage{adjustbox}
\usepackage{lipsum}
\usepackage[breaklinks]{hyperref}
\usepackage{enumitem}
\usepackage{listings}
\usepackage{xcolor}
\usepackage{tikz}
\usepackage{fontawesome5}
\usepackage[T1]{fontenc}
\usepackage[margin=1in]{geometry}

\usepackage[T1]{fontenc}
\usepackage[scaled=0.85]{DejaVuSansMono}

\usepackage[most]{tcolorbox}
\usepackage{varwidth}

\usepackage{url}

\definecolor{mygreen}{RGB}{220,255,220}
\definecolor{myred}{RGB}{255,220,220}
\definecolor{myblue}{RGB}{220,235,255}

\newcommand{\good}[1]{\colorbox{mygreen}{#1}}
\newcommand{\bad}[1]{\colorbox{myred}{#1}}
\newcommand{\note}[1]{\colorbox{myblue}{#1}}
\definecolor{headerblue}{HTML}{D2E3FC}
\definecolor{headerlavender}{HTML}{E5D7E9}
\definecolor{headergreen}{HTML}{d4e7d0}
\definecolor{headerpeach}{HTML}{F3D9C9}
\newcommand{\hr}{\par\medskip\noindent\rule{\linewidth}{0.4pt}\par\medskip}

\lstdefinestyle{prompt}{
    basicstyle=\ttfamily\footnotesize,
    breaklines=true,
    columns=fullflexible,
    keepspaces=true,
    frame=single,
    framesep=5pt,
    framexleftmargin=5pt,
    xleftmargin=12pt,
    xrightmargin=6pt,
    rulecolor=\color{black},
    backgroundcolor=\color{white},
    showstringspaces=false,
    upquote=true,
    aboveskip=10pt,
    belowskip=10pt,
    literate={-}{{-}}1
}
\tcbset{
    examplebox/.style={
        enhanced,
        colback=white,
        colframe=gray,
        boxrule=0.6pt,
        arc=2mm,
        left=2mm,
        right=2mm,
        top=1mm,
        bottom=1mm,
        fonttitle=\bfseries,
        coltitle=black,
        title filled=false,
    }
}

\title{Evaluating Memory Structure in LLM Agents}

\author{
Alina Shutova$^{\,*\,\dagger}$\hspace{19px}
Alexandra Olenina$^{\,\dagger\,\ddag}$\hspace{19px}
Ivan Vinogradov$^{\,\dagger\,\ddag\,\star}$\hspace{19px}
Anton Sinitsin$^{\,\dagger}$ \\
}

\begin{document}

\maketitle

\begin{abstract}

Modern LLM-based agents and chat assistants rely on long-term memory frameworks to store reusable knowledge, recall user preferences, and augment reasoning.
As researchers create more complex memory architectures, it becomes increasingly difficult to analyze their capabilities and guide future memory designs.
Most long-term memory benchmarks focus on simple fact retention, multi-hop recall, and time-based changes.
While undoubtedly important, these capabilities can often be achieved with simple retrieval-augmented LLMs and do not test complex memory hierarchies.
To bridge this gap, we propose StructMemEval --- a benchmark that tests the agent's ability to \underline{organize its long-term memory}, not just factual recall.
We gather a suite of tasks that humans solve by organizing their knowledge in a specific structure: transaction ledgers, to-do lists, trees and others.
Our initial experiments show that simple retrieval-augmented LLMs struggle with these tasks, whereas memory agents can reliably solve them if prompted how to organize their memory. However, we also find that modern LLMs do not always recognize the memory structure when not prompted to do so. This highlights an important direction for future improvements in both LLM training and memory frameworks.


\end{abstract}

\begingroup
\makeatletter
\renewcommand\@makefnmark{}
\footnotetext{\hspace{-15px}$^*$HSE University,
$^\dagger$Yandex,
$^\ddag$YSDA,
$^\star$New Economic School.
Correspondence to: \texttt{ant.sinitsin@gmail.com}\,.}
\makeatother
\endgroup


\section{Introduction}\label{sect:intro}

As Large Language Models (LLMs) keep improving, we trust them with increasingly longer and more difficult tasks~\citep{metr}. Frontier models assist users over long stretches of time~\citep{xu-etal-2022-beyond,jang-etal-2023-conversation,wang2025recursively}\nocite{zhang-etal-2023-mind,zhang-etal-2025-bridging,user_preferences_with_memory}, work with large documents~\citep{memgpt,chevalier2023adapting}, perform long-form reasoning~\citep{openai_arc_prize_o3,phan2025humanitysexam} and agentic tasks~\citep{erdogan2025plan,merrill2026terminalbenchbenchmarkingagentshard}\nocite{hu2026memoryageaiagents}.
To solve these tasks, a model needs to process vast amounts of information, both input sources and its own workings. As LLMs deal with harder problems, it becomes infeasible to keep all that information in their working memory.

Recent works circumvent this by equipping the LLM with long-term memory: allowing the agent to offload its knowledge from working memory to an external database and retrieve it only when necessary~\citep{memgpt,chatgpt_memory}.
Memory designs vary between methods: from simple look-up dictionaries~\citep{memgpt,memoryllm} to complex frameworks~\citep{mem0,zep} and entire operating systems~\citep{memos,memoryos}.

To compare between memory architectures, researchers and practitioners evaluate them on specialized benchmarks such as LOCOMO~\citep{locomo} and LongMemEval~\citep{longmemeval}.
These benchmarks typically let the LLM process a scenario (e.g. personal assistant, corporate, agent) and form memories, then ask questions to test its recall: single-hop, multi-hop, time-based queries, and others.
However, recent works have shown that LLMs do not necessarily need complex memory architecture to answer these kinds of questions~\citep{emem,memorybench}. In one such work,
\citet{emem} demonstrate that a simple retrieval system (EMem) can outperform sophisticated memory architectures on both LOCOMO and LongMemEval. This leads to a question: what kinds of problems \textit{do} require complex memory hierarchies? What capabilities do they add?

In this work, we focus on one answer: that \textit{memory allows LLM agents to organize their knowledge}. Instead of retrieving messages as-is, memory-augmented LLMs can extract knowledge and assemble it into a structure that best fits their task, such as arranging knowledge into graphs, maintaining todo lists, sorting events by category, calculating running statistics, and more.
If a problem requires knowledge organization, simple message retrieval will not be able to solve it past certain size, but a more flexible agentic memory may succeed if the controlling LLM is capable enough.

To test this hypothesis, we gather a suite of tasks that a ``notepad-augmented human'' would solve by maintaining ledgers, drawing graphs, tracking states and similar.
We find that even simple tasks of this kind prove difficult for retrieval-augmented agents, but proper memory-augmented agents fare significantly better.
When we explicitly prompt LLM agents to organize their memory, they can reliably solve tasks on a scale that retrieval-augmented LLMs struggle with, but perform less reliably without ``hints''.
Informally, LLMs that can solve abstract algorithmic tasks often fail to recognize the same structure patterns in the wild.
Our main contributions can be summarized as follows:\begin{enumerate}[leftmargin=*]
    \vspace{-7px}\item We gather StructMemEval, a suite of memory evaluation scenarios that focus on knowledge organization: tracking financial settlements, reasoning about family trees and organizations, tracking changing states from indirect observations, analyzing user preferences from events.
    \vspace{-3px}\item We analyze the effectiveness of different retrieval and memory-based LLM agents on memory organization tasks. Our findings suggest that retrieval-only systems fail after a certain problem size, whereas memory agents can maintain effectiveness longer, depending on the model.
    \vspace{-3px}\item We systematically analyze the failure modes of memory agents on StructMemEval tasks and report main failure modes: incorrect structures, memory hallucinations, record duplication, and others.
    \vspace{-3px}\item We release \footnote{Dataset and code are available at https://github.com/yandex-research/StructMemEval.} the main set of 51 scenarios and ${\ge}200$ additional tasks for analysis. Our scenarios decouple memory organization from other LLM capabilities to help inform future memory designs.
\end{enumerate}

\section{Background}\label{sect:background}

\textbf{Transformer Working Memory.} Transformer LLMs normally store task-specific information in the form of a Key-Value (KV) cache. This cache contains token-level vector representations used by attention layers on every inference step~\citep{transformer}. For modern LLMs, this cache is limited to few ${\times} 10^{4-5}$ tokens with a few extreme cases going over a million tokens~\citep{glm2024chatglmfamilylargelanguage,yang2025qwen251mtechnicalreport}. While this limit can be extended~\citep{yarn,longchat,pekelis2024llama3gradient}, a large KV cache fills accelerator memory\footnote{E.g. storing 1M tokens for \href{https://huggingface.co/Qwen/Qwen2.5-7B-Instruct-1M}{Qwen2.5-7B-Instruct-1M} in \texttt{bfloat16} takes up over 180GiB GPU memory.} and slows down inference.

If the LLM needs to process more information than its KV cache allows, there are two main strategies for achieving this: compressing the KV cache or using external long-term memory. The former includes KV quantization~\citep{kivi,kvquant,aquakv}, pruning~\citep{h2o,streamingllm}, offloading techniques~\citep{infinigen} and other techniques that reduce the overhead from having a large KV cache. The latter bypasses the need for large KV cache by allowing the LLM to offload its knowledge to an external database and retrieve it later.\nocite{aquakv}

\textbf{Memory-Augmented LLMs.} Even before the LLMs, NLP researchers have allowed their models to retrieve relevant knowledge from corpora to give it access to additional knowledge and improve factuality~\citep{realm,dense_passage_retrieval,rag}\nocite{rel_rnn_memory_attn}.
However, not all problems can be solved with a static database: for instance, when dealing with user preferences or working memory, the model must be able to update its knowledge to maintain relevant information~\citep{memformer}.
To address this, \citet{memgpt} propose MemGPT --- an external memory system that equips LLM agents with tools to create and update memory entries for later retrieval.

Since then, multiple lines of work propose extensions to this idea using knowledge graphs~\citep{zep,mem0,QRMem}, hierarchy~\citep{hmem,CarMem,himem,memweaver} insights from cognitive science~\citep{nemori}, note taking~\citep{amem,mem-agent}, and others~\citep{cdmem,lightmem,memmachine,superlocalmemory,implicit_memory,hymem}.
Aside from storing user preferences, agentic memory is also used for knowledge updates~\citep{memoryllm,mplus}, iterative ``evolutionary'' problem solving~\citep{gepa,ace}, programming~\citep{researchcodeagent,talm}, embodied and virtual agents~\citep{llm-powered-embodied-agent-with-memory,robomemory,karma,memory_augmented_state_machine_prompting}. Recent works propose even more complex memory frameworks and operating systems~\citep{memos,memoryos} or fine-tuning the LLM to a specific memory type~\citep{memllm,memory-r1,memskill}.\nocite{huang2026rethinkingmemorymechanismsfoundation}

\textbf{Benchmarking Long-term Memory} frameworks is an inherently difficult task. Two of the most popular benchmarks for memory agents, LOCOMO~\citep{locomo} and LongMemEval~\citep{longmemeval}, focus on a chat assistant use case where the agent is expected to recall individual facts, do multi-hop look-ups and update its knowledge over time as the context changes. Other benchmarking scenarios are based on interactive conversations with user feedback~\citep{dialsim,storybench,memorybench} or convert existing long-context datasets~\citep{memoryagentbench}. Other more specialized benchmarks target test-time learning~\citep{evo-memory}, personalization and others~\citep{MIP-Bench, personamem-v2,locco-2025.findings-acl.1014}. However, recent works have found that many of these evaluation tasks do not use complex memory hierarchies~\citep{emem,memorybench,zep}. Notably,~\citet{emem} introduce EMem and Emem-G: simple retrieval baselines that outperform more complex memory structures on both LOCOMO and LongMemEval. We hypothesize that these benchmarks do not test the more advanced components of agentic memory, making it harder to analyze their effectiveness.\nocite{li2026locomoplusbeyondfactualcognitivememory}


\section{StructMemEval}\label{sect:method}
\vspace{-7px}

In this work, we create a benchmark for evaluating one specific capability of memory agents: how well can they structure their long-term memory for a given problem. The main difficulty for such a benchmark is that we cannot rely on a specific internal memory architecture. As we discussed in Section~\ref{sect:background}, modern memory frameworks use varied internal representations from look-up dictionaries, Zettelkasten notes, to graphs and temporal databases. Given the same task, Mem0-G and Zep will map the desired knowledge structure into a graph database, whereas A-Mem and MemAgent would use interlinked notes.
To compare different memory frameworks, we need our task suite to be implementation-agnostic. To that end, we only evaluate the final answers, not the internal structure, but we design our problems in such a way that solving them requires the agent to organize its memory.

Another important consideration is that our tasks need to benchmark the memory structure specifically. For instance, if we were to give the LLM an extremely difficult programming task with limited context window, the agent would have to use long-term memory. However, it would also require programming and design capabilities, which would contaminate analysis. If an agent failed to solve the problem, it would be unclear if the error was caused by ineffective memory organization or simple coding error. To decouple our benchmark from other capabilities, we deliberately select tasks that are simple given correct memory organization but nearly impossible without it. For convenience, we organize our benchmark around specific memory organization patterns that humans have already used: trees, ledgers, to-do lists, indexes and others. For each pattern, we gather problems where a human expert would follow that structure in their notes. We discuss each structure type below.

\begin{figure}[t]
    \centering
    \includegraphics[width=0.99\linewidth]{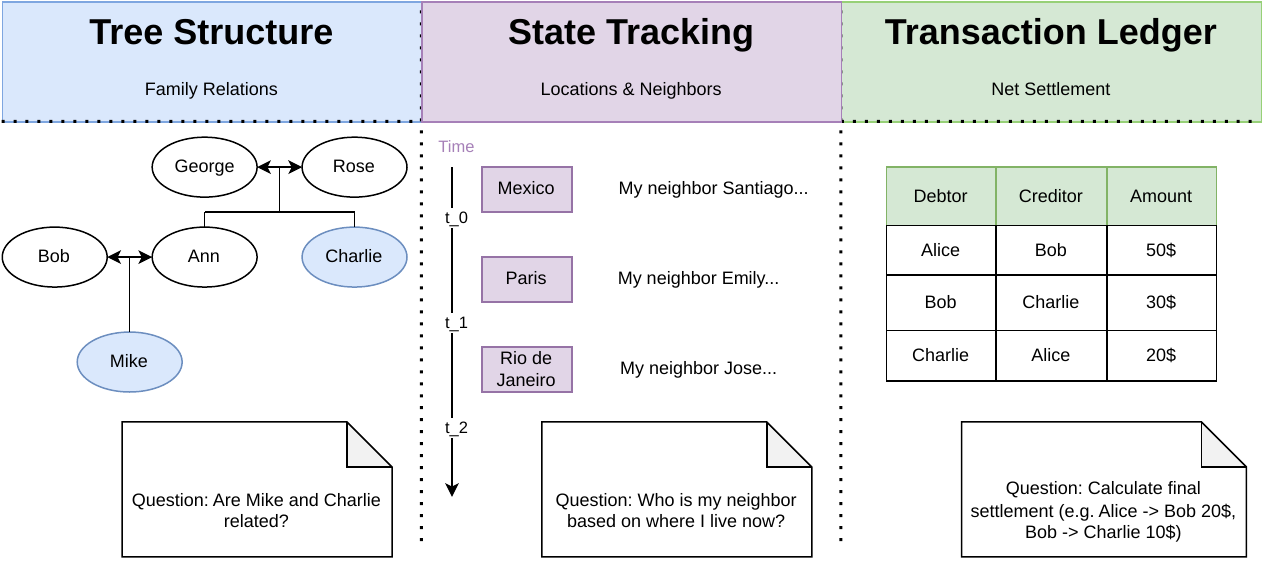}
    \vspace{-5px}
    \caption{Example memory organization problems from three of StructMemEval categories: tree-structured problems, state tracking and transaction ledgers. Crucially, the agent does not receive these structures directly and must construct them (or equivalents) from raw message or event history.}
    \label{fig:benchmark_summary}
    \vspace{-15px}
\end{figure}


\textbf{Tree-structured problems} require the agent to maintain hierarchical knowledge such as taxonomies, family trees, corporate hierarchies or codebases. To test this ability, we design problems based on family trees and corporate hierarchies. The memory agent is given a sequence of messages that encode relations (``A is B's stepdaughter'') and is expected to maintain the full graph. Some of the relations are implied (e.g. if ``C is B's wife'', then the last example also implies that ``C is A's parent''). Notably, we found that such indirect queries often confuse purely retrieval-based LLMs.

\textbf{State tracking problems} are tasks where one or multiple entities can change their state over time. A practical example is when someone moves cities, they are then no longer neighbors with those left behind. A similar state transition can happen with tasks on KANBAN boards or functions during codebase refactoring. We gather scenarios where individual messages only matter in the context of their state and keeping track of state transition is important for answering the question.
For instance, consider a scenario where 1) a user interacts with their neighbors, then 2) moves to a different location and interacts with a different group of neighbors there, before 3) returning to their original location. These scenarios can confuse retrieval-based systems because the use of ``neighbor'' is contextual. Simple retrieval often falsely includes neighbors from stage (2) even when the question explicitly specifies stage (3). In turn, a memory agent can track user's state and organize neighbors by location.

\textbf{Counting-based problems} are tasks that involve maintaining and reconciling totals. For example, an accountant may use offsetting or netting diagrams to determine final settlement amounts (i.e., which party owes how much), alongside other task-specific organization practices~\citep{kieso_intermediate_ifrs_4e}. Similar scenarios arise when analyzing network traffic, processing scientific observations or a patient's medical record.
These accounting problems can be seen as a special case of mathematical reasoning, except that the agent ``reasons'' by aggregating transactions in their long-term memory over hundreds of messages.
We gather tasks where the agent observes a history of transactions (e.g. ``A supplied \$X for B'') between multiple parties, and must then compute the final settlement after netting (i.e. canceling circular debts). These tasks also include messages unrelated to the settlement.

\textbf{Recommendation problems} are workloads where the LLM must organize and reason about user preferences from event history. We gather scenarios where a user consumes tv series, movies, art, board games, books, and music and provides feedback, including general opinion (e.g. ``rewatched X, still holds up perfectly'') and more specific preferences (``Y is only of a few directors who gets Z''). Then, the agent is tasked with questions: does the user like music of genre A more than B? What is the probability that they rate a thriller positively? Did they watch more movies of type X or Y? We deliberately choose questions that require aggregation and running statistics, not individual recall.

\textbf{Data curation \& Evaluation.} We gather scenarios using both human annotators and LLM augmentation with human verifiers.
Every scenario consists of a conversation history and a set of evaluation questions asked at different conversation depths: testing if two people are related for trees, evaluating the final debt settlement amount, etc.
Using synthetic scenarios allows us to release tasks themed after sensitive user and business applications without data privacy risks~\citep{overview_privacy_llm,overview_synthetic_data}.
We evaluate accuracy using LLM-as-a-judge~\citep{zheng2023llm-as-a-judge} \texttt{gpt-4o} with a prompt that focuses on factual correctness against the reference answer. 
We additionally performed manual verification of the results on the hardest subset of the benchmark, the count-based tasks.



Overall, we gathered 207 evaluation scenarios: 90 tree-based, 45 count-based tasks, 42 state tracking tasks and 30 recommendation problems, with over 2000 evaluation questions. Each subset contains problems of different size, between 10 and 500 messages. Appendix~\ref{app:details_data_collection} contains additional details on data gathering and curation for every subset.
For ease of use in future evaluations, we organize StructMemEval benchmark as follows:\begin{itemize}[leftmargin=*]
    \item \textbf{The main evaluation set consists of 51 problems:} 10 tree-based, 15 count-based, 14 state tracking and 12 recommendation scenarios, the longest ones in each category (${\ge}250$ messages). This is the main StructMemEval benchmark and we encourage future comparisons to use this problem set.
    \item \textbf{The extended set of 207 scenarios} is meant primarily for scaling analysis. In Section~\ref{sect:experiments}, we use the shorter problems to compare how different LLMs, memory, and retrieval systems compare across different problem lengths.
\end{itemize}


\textbf{Memory organization hints.} For each scenario, we provide an optional ``memory organization hint'': an informal text prompt that explains how a human would organize their knowledge for the task at hand. We use those hints for analysis only, \textit{the main StructMemEval setup is evaluating without hints}.
Instead, we use memory organization hints to diagnose error modes: if an agent fails to solve a given problem as is, but then solves it reliably with the hint, then its original mistake stemmed from poor knowledge structure. If, however, that agent still fails even with ``hints'', then the error is not due to wrong memory organization strategy, but in poor execution: failing to maintain the chosen memory structure or failing to properly utilize it when answering the test questions.

\vspace{-10px}
\section{Evaluations \& Analysis}\label{sect:experiments}
\vspace{-8px}

In this section, we analyze how retrieval and memory agents perform on memory organization tasks with different setups and LLM controllers. We organize our evaluations as follows:\begin{itemize}[leftmargin=*]
    \item In Section~\ref{sect:experiments_lengths}, we evaluate retrieval and memory agents for varying task lengths and verify that retrieval-based systems do not scale past certain StructMemEval problem sizes.
    \item Section~\ref{sect:experiments_main_results} evaluates mem-agent, \texttt{Mem0}, basic retrieval, and EMem on the main StructMemEval benchmark and compares the effect of different LLM backbones.
    \item Section~\ref{sect:experiments_hints} uses memory organization hints to decouple the choice of memory structure from the LLM's ability to follow that structure across long evaluation scenarios.
    \item Section~\ref{sect:experiments_error_analysis} is a detailed error analysis: we report specific agent behaviors that trigger failure modes, compare backbone LLMs and analyze the impact of different prompting strategies.
\end{itemize}

\vspace{-5px}
\subsection{Retrieval and Memory at Different Problem Sizes}\label{sect:experiments_lengths}
\vspace{-5px}


\begin{figure}[t]
    \centering
    \includegraphics[width=0.46\linewidth,trim=10px 20px 20px 20px,clip]{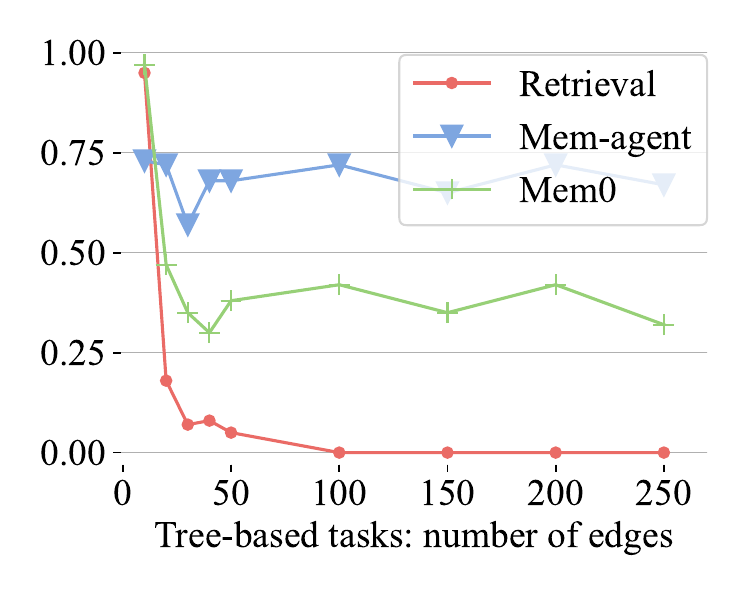}
    \includegraphics[width=0.46\linewidth,trim=10px 20px 20px 20px,clip]{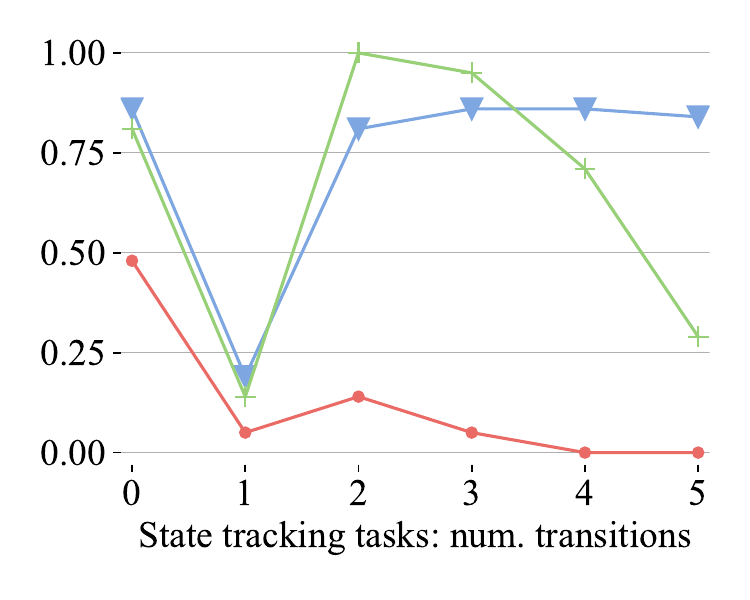}

    \caption{Analyzing retrieval and memory agents with \texttt{gemini-3.1-pro} on different scenario lengths for \textbf{(left)} tree-based and \textbf{(right)} state tracking problem sets. See detailed setup in Section~\ref{sect:experiments_lengths}.}
    \label{fig:4.1_length_treebased_and_state}
    \vspace{-10px}
\end{figure}

\begin{wrapfigure}{r}{0.45\textwidth}
  \centering
  \includegraphics[width=\linewidth,trim=10px 20px 20px 20px,clip]{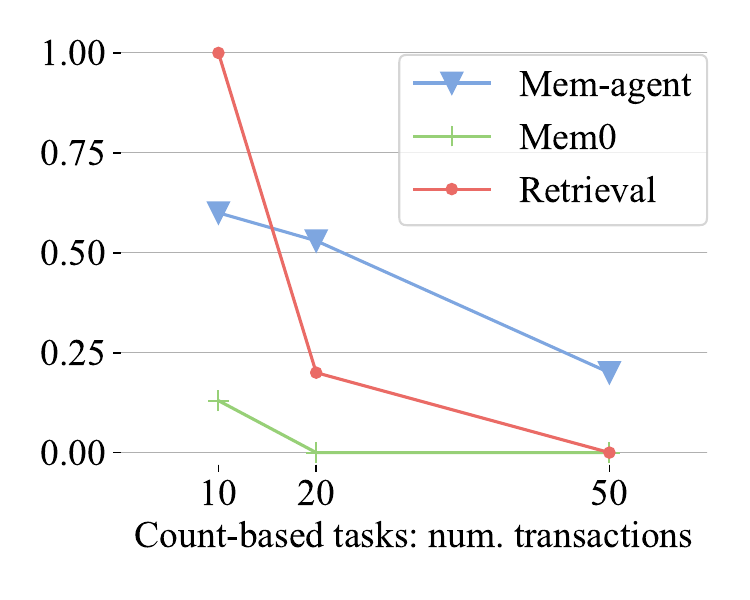}
  \vspace{-15px}
  \caption{Retrieval and memory agents on count-based tasks, \texttt{gemini-3.1-pro}, see Section~\ref{sect:experiments_lengths}.}
  \label{fig:4.1_accounting}
  \vspace{-45px}
\end{wrapfigure}

Before comparing more advanced systems, we need to verify the main assumption of Section~\ref{sect:method}: that StructMemEval \textit{requires} memory organization, not just factual retrieval. To test this, we evaluate a retrieval-only agent that, by definition, cannot organize its memory, across different problem sizes.
More specifically, we evaluate the following systems:\begin{enumerate}[leftmargin=*]
    \vspace{-5px}\item \textbf{Retrieval:} we use the retrieval-only baseline from \texttt{Mem0} codebase~\citep{mem0} using OpenAI \texttt{text-embedding-3-large}~\citep{openai_embeddings_api} with top-10 passage retrieval.
    \vspace{-3px}\item \textbf{Mem-agent:} markdown-based memory from Mem-agent~\citep{mem-agent}.
    \vspace{-3px}\item \textbf{Mem0:} a popular agentic memory architecture proposed by~\citet{mem0}.
\end{enumerate}\vspace{-5px}

We use \texttt{gemini-3.1-pro}~\citep{gemini31pro_blog_2026} for all main experiments. Appendix~\ref{app:details_memory_frameworks} contains additional details and configurations for retrieval and each memory framework.
We evaluate memory frameworks with default hyperparameters.
Figure~\ref{fig:4.1_length_treebased_and_state} (left) reports graph-based problems, where the scenario length is measured as the number of graph edges. Likewise, Figure~\ref{fig:4.1_length_treebased_and_state} (right) contains state tracking problems, ordered by the total number of state transitions. For accounting problems in Figure~\ref{fig:4.1_accounting}, we evaluate accuracy based on the total number of transactions in message history. Distractor messages do not count towards task complexity.
Across all scenario types, retrieval-only LLMs can reliably solve small tasks, but quickly fall off once the task complexity no longer fits in the retrieval window. In turn, both memory agents start off worse, but scale better into more complex tasks.
We report additional models-scenario pairs and top-$k$ configurations in Appendix~\ref{app:alternative_models} and provide error analysis in Section~\ref{sect:experiments_error_analysis}.

\begin{table*}[b]
\vspace{-16px}
\centering
\setlength{\tabcolsep}{10pt}
\renewcommand{\arraystretch}{1.1}
\caption{Evaluation of different long-term memory and retrieval agents on StructMemEval (main set, without hints). The entries are accuracy scores (higher is better), ``Total'' is the equal-weighted subset average. In this table, all methods use \texttt{gemini-3.1-pro} as the backbone LLM for fair comparison.}
\label{tab:4.2_main_frameworks}
\vspace{-8px}
\begin{tabular}{l|cccc|c}
\toprule
\textbf{Agent Type} & \textbf{State-tracking} & \textbf{Tree-based} & \textbf{Count-based} & \textbf{Recsys} & \textbf{Total} \\
\midrule
Retrieval & 0.00 &  0.00&  0.00&  0.22 & 0.06 \\
EMem & 0.29 & 0.57 & 0.00 & 0.14 & 0.25 \\
EMem-G & 0.36 & 0.55 & 0.00 & 0.14 & 0.26 \\
Mem-agent & 0.84 & 0.98 & 0.16 & 0.37 & 0.59 \\
\texttt{Mem0} & 0.29 & 0.72 & 0.00 & 0.18 & 0.30 \\
\texttt{Mem0-G} & 0.00 & 0.72 & 0.00 & 0.17 & 0.22 \\
\bottomrule
\end{tabular}
\vspace{-12px}
\end{table*}

\subsection{Evaluating Memory Architectures and LLM Controllers}\label{sect:experiments_main_results}

Next, we analyze how memory organization frameworks vary across different memory frameworks and backbone LLMs using the main StructMemEval setup (51 hard problems). We use the more recent 
\texttt{gemini-3.1-pro}~\citep{gemini31pro_blog_2026} backbone and evaluate several agent configurations:\begin{itemize}[leftmargin=*]
    \vspace{-5px}\item \textbf{Retrieval:} chat history retrieval using \texttt{gemini-3.1-pro} controller and \texttt{text-embedding-3-large} embeddings, all other parameters are equal to the Section~\ref{sect:experiments_lengths} setup.
    \vspace{-3px}\item \textbf{EMem \& EMem-G:} improved retrieval agents~\citep{emem} that convert chat history into 
    Elementary Discourse Units (EDUs) for retrieval, with all recommended parameters (\texttt{text-embedding-3-large}, linking\_top\_k=30, qa\_top\_k=10), \texttt{gemini-3.1-pro} LLM.
    \vspace{-3px}\item \textbf{Mem-agent:} markdown-based memory agent from~\citet{mem-agent}, same as in the previous section.
    \vspace{-3px}\item \textbf{Mem0:} the same memory agent from~\citet{mem0}.
\end{itemize}\vspace{-5px}

Table~\ref{tab:4.2_main_frameworks} summarizes our evaluations. The retrieval-only agent consistently fails to solve memory organization problems, which is expected and aligns\footnote{Note that Table~\ref{tab:4.2_main_frameworks} uses the main StructMemEval that consists of the longest problems in each category.} with our earlier observations in Section~\ref{sect:experiments_lengths}. One notable exception is recommender tasks, where it gets a better-than-random estimate at multiple choice questions (e.g. which movies are viewed more often, A or B). Furthermore, retrieval-only agent retains poor accuracy even if we increase the search budget to top-20. EMem retrieval makes better use of its budget and scores higher, but still consistently below mem-agent. In turn, the two agentic memory frameworks (\texttt{Mem0} and Mem-agent) perform well.

Next, we measure the impact of the backbone LLMs on knowledge organization. We use Mem-agent memory with multiple LLM backbones of different sizes: \texttt{gemini-3.0-flash}~\citep{gemini_3_flash_2025}, \texttt{gemini-3.1-pro}, \texttt{gpt-5.5}~\citep{singh2025gpt5systemcard}, \texttt{deepseek-v4-flash} and \texttt{deepseek-v4-pro}~\citep{deepseek_v4_2026} (the latter
two are open-weight). The results in Table~\ref{tab:4.2_main_models} and Table~\ref{tab:4.2_extra_models} indicate that all backbone models can, to some extent, organize their memory and outperform the retrieval baselines. However, individual subset performance varies significantly between LLMs: for instance, the ``flash'' models perform poorly on accounting tasks. We evaluate all models except Opus 4.7, gpt-5.5, claude sonnet 5 and gemini-3.8-flash with 5 seeds and report adjusted standard deviations, using bootstrapping for the average score. We report additional memory and retrieval setups in Appendix~\ref{app:alternative_models}. In the next two sections, we delve deeper into how different LLMs fail to organize their knowledge and whether it can be alleviated.

\begin{table*}[!h]
\centering
\setlength{\tabcolsep}{9pt}
\renewcommand{\arraystretch}{1.3}
\caption{Evaluation of mem-agent with different backbone LLMs on StructMemEval (main set, without hints). The entries are accuracy scores (higher is better), ``Total'' is the equal-weighted average of 4 subsets.} 
\label{tab:4.2_main_models}
\begin{tabular}{l|cccc|c}
\toprule
\textbf{Backbone LLM} & \textbf{State-tracking} & \textbf{Tree-based} & \textbf{Count-based} & \textbf{Recsys} & \textbf{Total} \\
\midrule
\texttt{gemini-3.0-flash} & 0.66 $\pm$ 0.05& 0.98 $\pm$ 0.002& 0.00 $\pm$ 0.00& 0.31 $\pm$ 0.01& 0.49  $\pm$  0.01\\
\texttt{gemini-3.1-pro} & 0.84 $\pm$ 0.03& 0.98 $\pm$ 0.003& 0.16 $\pm$ 0.02 & 0.37 $\pm$ 0.01 & 0.59 $\pm$ 0.02 \\
\texttt{deepseek-v4-flash} & 0.56 $\pm$ 0.02 & 0.81 $\pm$ 0.02 & 0.00 $\pm$ 0.00 & 0.13 $\pm$ 0.01 & 0.38 $\pm$ 0.01\\
\texttt{deepseek-v4-pro} & 0.51 $\pm$ 0.04 & 0.78 $\pm$ 0.03 & 0.01 $\pm$ 0.01 & 0.24 $\pm$ 0.01 & 0.39 $\pm$ 0.02\\
\texttt{gemini-3.5-flash} & 0.77 $\pm$ 0.04 & 0.74 $\pm$ 0.05 & 0.04 $\pm$ 0.02 & 0.12 $\pm$ 0.02 & 0.42 $\pm$ 0.02\\
\bottomrule
\end{tabular}
\end{table*}

\begin{table*}[!h]
\centering
\setlength{\tabcolsep}{14pt}
\renewcommand{\arraystretch}{1.3}
\caption{Evaluation of mem-agent with newer backbone LLMs on StructMemEval (main set, without hints). The entries are accuracy scores (higher is better), ``Total'' is the equal-weighted average of 4 subsets.} 
\label{tab:4.2_extra_models}
\begin{tabular}{l|cccc|c}
\toprule
\textbf{Backbone LLM} & \textbf{State-tracking} & \textbf{Tree-based} & \textbf{Count-based} & \textbf{Recsys} & \textbf{Total} \\
\midrule
\texttt{claude-opus-4.7} & 0.79 & 0.30 & 0.00 & 0.00 & 0.27 \\
\texttt{gpt-5.5} & 0.57 & 0.50 & 0.00 & 0.33 & 0.35 \\
\texttt{claude-sonnet-5} & 0.71 & 0.90 & 0.00 & 0.37 & 0.50 \\
\texttt{gemini-3.8-flash} & 0.64 & 0.92 & 0.07 & 0.27 & 0.47 \\
\bottomrule
\end{tabular}
\end{table*}

\subsection{Memory Organization Hints}\label{sect:experiments_hints}

\begin{figure}[b]
    \centering
    \vspace{-10px}
    \includegraphics[width=\linewidth,trim=0px 0px 0px 0px,clip]{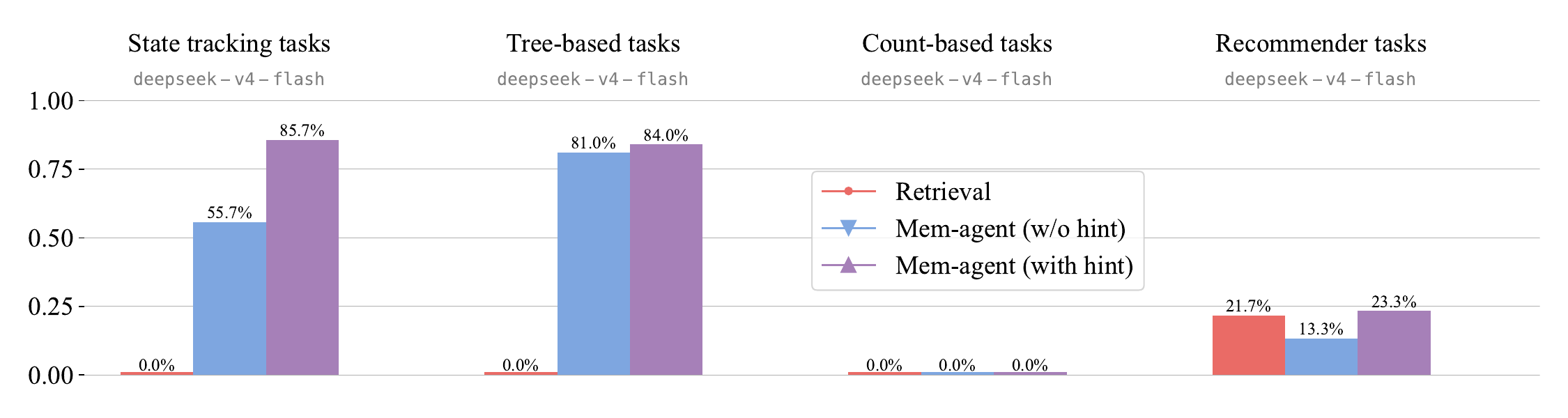}
    \caption{Evaluation of Mem-agent with and without hints across StructMemEval (main) subsets with \texttt{deepseek-v4-flash}. For this LLM, count-based tasks have zero accuracy even with hints.}
    \label{fig:4.3_hints_deepseek}
\end{figure}

While memory agents outperform retrieval-augmented LLMs in our evaluations, they are still far from perfect accuracy, particularly for smaller models such as \texttt{deepseek-v4-flash} and \texttt{gemini-3.0-flash}. There are two possible explanations: either the agents fail to organize their knowledge effectively, or they ``do not try''. The former seems counterintuitive because LLMs are known to solve complex programming tasks using similar data structures: trees, heaps, state machines~\citep{chen2021codex,jain2025livecodebench}. However, recent studies have shown that language models often fail to solve reasoning problems in unfamiliar contexts ~\citep{laban2025llms,du2025contextlengthhurtsllm}\nocite{cohen2025forget,rodionov2026reasoningshiftcontextsilently,devbunova2026evaluation,needham2025large,truong2025persona,abedin2025arithmattack}.
In other words:\begin{itemize}[leftmargin=*]
    \vspace{-5px}\item \textbf{Option A:} the agent fails because it cannot follow the knowledge organization pattern required by the task. This could be caused by insufficient reasoning capability, hallucinations, or failure to properly utilize the memory framework.
    \vspace{-3px}\item \textbf{Option B:} the agent could solve the task correctly if prompted to organize its memory, but it does not recognize that the problem needs memory organization.
\end{itemize}\vspace{-5px}

To distinguish between the two possibilities, we manually prompt the LLMs to follow a task-specific memory structure chosen by a human (see hint texts in Appendix~\ref{app:memory_organization_hints_prompts}). Intuitively, if the agent still fails to solve the problem when given detailed instructions, it would mean the LLM fundamentally cannot understand or follow the structure. However, if the instruction helps, it means the LLM ``knows the algorithm'' for knowledge organization, but didn't recall it unprompted.

\begin{wrapfigure}{r}{0.4\textwidth}
    \vspace{-22px}
    \centering
    \includegraphics[width= \linewidth,trim=0px 0px 0px 0px,clip]{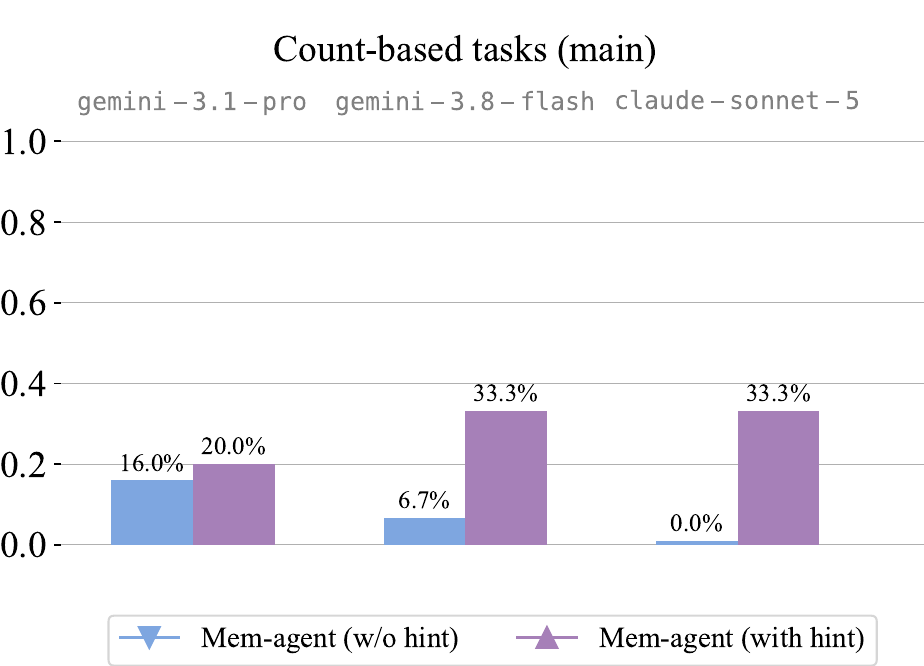}
    \vspace{-15px}
    \caption{Evaluation of \texttt{gemini-3.1-pro}, \texttt{gemini-3.8-flash}  and \texttt{claude-sonnet-5} with and without hints on count-based tasks.} 
\label{fig:4.3_hints_gemini_and_mem0agent}
\vspace{-25px}
\end{wrapfigure}

The results in Figure~\ref{fig:4.3_hints_deepseek} demonstrate that adding hints can significantly improve the agent's ability to solve StructMemEval tasks, though they still do not achieve perfect accuracy. Notably, \texttt{deepseek-v4-flash} does not solve count-based (accounting) problems even after detailed prompts. In Figure~\ref{fig:4.3_hints_gemini_and_mem0agent} (left), we apply the same accounting hints to \texttt{gemini-3.1-pro} where hints do increase accuracy. We also apply similar hints to \texttt{Mem0} agent, observing similar improvements. Curiously, the gap between memory agents with and without hints is larger than the difference between \texttt{Mem0} and Mem-agent. In summary, memory organization hints \text{do} improve memory agents' performance, but they do not ``solve'' the benchmark, which implies there are failure modes not covered by the hint. The next section analyzes this further.

\subsection{Error Analysis with Memory Organization Hints}\label{sect:experiments_error_analysis}

\begin{figure}[t]
    \centering
    \begin{minipage}[t]{0.485\linewidth}
    \begin{tcolorbox}[
    examplebox, colframe=headerblue,
    title={Bidirectional Link Error, Task 7},
    ]
    \footnotesize
    \textbf{Input message:} 
    \vspace{0.5mm}\\
    \ttfamily
    - Johnny works with Diane \\
    - Johnny works with Sam \\
    \normalfont
    \textbf{Question:} 
    \vspace{0.5mm}\\
    \ttfamily
    - Are Diane and Sam related through colleagues relations?
    \vspace{0.5mm}\\
    \normalfont
    \textbf{Memory state:}
    \vspace{3.5mm}\\
    \faFile\;\tt{johny.md:\{Colleagues: Diane, Sam\}} \\
    \faFile\;\tt{diane.md: absent file}\\
    \normalfont
    \textbf{Answer:} 
    \vspace{1.7mm}\\
    \ttfamily
    \bad{- I have no records about Diane.}
    \vspace{0.5mm}\\
    \note{$\uparrow$ The model records one direction of} \\
    \note{the edge but forgets the reverse link.}
    \end{tcolorbox}
    \end{minipage}
    \hspace{-3px}
    \begin{minipage}[t]{0.51\linewidth}
    \begin{tcolorbox}[examplebox, colframe=headerblue,
                      title={Hallucinated Job Info, Task 1}]
    \footnotesize
    \textbf{Input message:}\\
    \ttfamily
    - Gregory works with Jose.
    
    \normalfont
    \textbf{Question:}\\
    \ttfamily
    - Are Gregory and Austin related through colleagues relations?\\
    \normalfont
    \textbf{Answer:}\\  
    \ttfamily{
    - Gregory is currently a \bad{Senior} \\
    \bad{Developer at Tech Solutions Inc.} and a colleague of \bad{Sarah}. \bad{Sarah} is the common link; she is a former colleague of Austin from their time together... \\
    \note{$\uparrow$ There is no person named Sarah in the} \\
    \note{dataset; job title, company name are}
    \note{also generated.}}
    \end{tcolorbox}
    \end{minipage}
    \caption{Examples of typical failures on tree-based tasks,  Mem-agent, \texttt{deepseek-v4-flash}.}
    \label{fig:4.4_tree_based_examples}
    \vspace{-15px}
\end{figure}

\begin{figure}[b]
    \vspace{-10px}
    \centering
    \begin{minipage}[t]{0.495\linewidth}
    \begin{tcolorbox}[
    examplebox, colframe=headerlavender,
    title={No Organization Hint, Task 4tr\_002},
    ]
    \footnotesize
    \ttfamily
    \normalfont
    \textbf{Memory state:}
    \vspace{1mm}
    \ttfamily\\
    \faFile\;\tt{user.md:\{Location: Ubud, Bali, \dots\}}\\
    \faFile\;\tt{franz.md:\{Pianist, Neighbor,\dots\}}\\
    \faFile\;\tt{marie.md:\{Filmmaker, Neighbor, \dots\}}\\
    \faFile\;\tt{elif.md: \{Teacher, Neighbor, \dots\}}
    
    \normalfont
    \vspace{7px}
    \textbf{Input:}  I'm moving to Montreal.
    \vspace{2mm}

    \textbf{Memory update:} 

    \good{\tt{user.md} $\leftarrow$ \{Location:\hspace{-3px} Montreal\}}
    
    \vspace{3mm}
    \textbf{Question:} Who is my neighbor now?
    \bad{Franz, Marie, Elif are all counted as ``neighbors''.} \\
    \note{The memory structure requires multiple updates.}
    \normalfont
    \end{tcolorbox}
    \end{minipage}
    \hfill
    \begin{minipage}[t]{0.495\linewidth}
    \begin{tcolorbox}[examplebox, colframe=headerlavender,
                      title={With Organization Hint, Task 4tr\_002}]
    \footnotesize
    \ttfamily
    \normalfont
        \textbf{Memory state:}
    \vspace{1mm}
    \ttfamily\\
    \faFile\;\tt{user.md:}
    
    \hspace{20px}\tt{\{Location: Ubud, Bali;}
    
    \hspace{20px}\tt{Bali neighbors: Franz, Elif;}
    
    \hspace{20px}\tt{Montreal neighbors: Marie, \dots\}}\\
    \faFile\;\tt{franz.md: \{Pianist, \dots\}}\\
    \faFile\;\tt{marie.md: \{Filmmaker, \dots\}}\\
    \faFile\;\tt{elif.md: \{Teacher,\dots\}}
    
    \normalfont
    \vspace{3px}
    \textbf{Input:}  I'm moving to Montreal.

    \textbf{Memory update:} 

    \good{\tt{user.md} $\leftarrow$ \{Location:\hspace{-3px} Montreal\}}\\
    \textbf{Question:} Who is my neighbor now?
    \good{Marie.} \note{\tt{No further updates required.}}
    \normalfont 

\end{tcolorbox}
    \end{minipage}
    \caption{Example memory structure on state tracking tasks and how the organization hint affects it.}
    \label{fig:4.4_state_tracking_examples}
\end{figure}

\begin{figure}[t]
    \centering
    \begin{minipage}[t]{0.49\linewidth}
    \begin{tcolorbox}[
    examplebox, colframe=headergreen,
    title={Record Duplication, Task 10.2},
    ]
    \footnotesize
    \textbf{Memory state:} \texttt{user.md} (excerpt)
    \vspace{0.5mm}\\
    \ttfamily
    |Payer|Paid for|Amount|Notes|\\
    |---------|----------------|-----------|---------|\\
    |Alice|All|180.00 EUR|train tickets|\\
    \bad{|Bob|All|18.00 EUR|drinks|}\\
    \bad{|Bob|All|18.00 EUR|beers at the pub|}\\
    \vspace{1.mm}\\
    \note{$\uparrow$ These two rows are duplicates:} \\
    \note{they describe the same 18.00 EUR}\\
    \note{expense paid by Bob.}\\
    \vspace{1.2mm}\\
    \dots\\
    \vspace{2.2mm}\\
    \bad{|Bob|Alice, Bob|15.00 EUR|parking|}\\
    \bad{|Bob|All|15.00 EUR|parking|}\\
    \note{$\uparrow$ Another duplicate. The rows also}\\
    \note{disagree about who benefited.}\\
    \end{tcolorbox}
    \end{minipage}
    \hspace{-3px}
    \begin{minipage}[t]{0.5\linewidth}
    \begin{tcolorbox}[examplebox, colframe=headergreen,
                      title={Spurious Records, Task 50.1 \& 10.2}]
    \footnotesize
    \textbf{Memory state:} \texttt{user.md} (excerpt)
    \vspace{0.5mm}\\
    {\ttfamily
    |Payer|Paid for|Amount|Notes|\\
    |---------|----------------|-----------|---------|\\
    \bad{|Alice|Bob|15 EUR|sandwich|} \\
    \vspace{1.2mm}\\
    \note{$\uparrow$ This row is entirely hallucinated:} 
    \note{the word ``sandwich'' does not appear}
    \note{anywhere in the dataset, and no}
    \note{combination of payer Alice and amount} 
    \note{15 EUR exists in this task.}}
    \vspace{2.2mm}\\
    \dots \\
    \vspace{2.2mm}\\
    \ttfamily{
\bad{|Alice|All|48 EUR|museum entry|}}

    \textbf{Original messages:} \\
    4)Alice: Museum -- €45 total for all.\\
    47)Bob: Museum entry was €48 total for three of us.\\
\note{$\uparrow$ Museum transactions merged into one.}
    \end{tcolorbox}
    \end{minipage}
    \caption{Example hallucinations on accounting tasks: duplication and spurious records.}
    \label{fig:4.4_accounting_examples}
    \vspace{-15px}
\end{figure}

To better understand why memory agents fail at StructMemEval problems, we examine how they organize their memory with and without hints. We gather Mem-agent trajectories
from Section~\ref{sect:experiments_hints} using \texttt{deepseek-v4-flash} LLM and review 1) the memory structure and update history and 2) how the agent uses its memory when answering questions.
We analyze these trajectories in two stages:
\begin{itemize}[leftmargin=*]
    \item \textbf{LLM-assisted exploration:} We gather pairs of trajectories for the same scenario \textit{with and without hints}. We then prompt an LLM assistant (we used \texttt{anthropic/claude-sonnet-4.6},~\citet{anthropic2026claudesonnet46}) to process one hint/no-hint pair (with identical messages) at a time and tally differences in agent behavior and memory state.
    \item \textbf{Manual verification:} We manually review LLM summaries and  verify that the divergences are genuine and look for repeating failure modes. We summarize our findings below.
\end{itemize}

\textbf{Case study 1: Tree-based tasks.} Without hint, Mem-agent partitions graphs (family or corporate) differently for each sample. Some nodes (i.e. people) are stored as a separate file while others are concatenated together. The most frequent error for no-hint agent is \textit{failing to account for bidirectional links}, i.e. if the model learns that A is linked to B (e.g. mother), it sometimes fails to also record that B is related to A (e.g. daughter), then fails to find the reverse path. Figure~\ref{fig:4.4_tree_based_examples} depicts one excerpt where this happens. The hint specifically asks the model to store graph vertices separately and account for both sides of every link, which leads to significantly higher accuracy in our previous experiments. The second most likely cause of errors is hallucinating nonexistent links when writing into memory.

\textbf{Case study 2: State tracking tasks.} Without hints, the agent often stores connected entries as separate records. For instance, the agent maintains the ``user'' entry and several separate files for each of the user's neighbors. When the user moves to a different location, the agent updates the user's own entry, but fails to update other persons' entries to mark them as connected to the previous location. Figure~\ref{fig:4.4_state_tracking_examples} illustrates this failure mode and how it changes with memory organization hint. Additionally, the agent sometimes fails to record state changes despite acknowledging them in conversation.

\textbf{Case study 3: Count-based tasks.} This case study is different in two regards. First, even high-capability models such as \texttt{gemini-3.1-pro} have poor accuracy on count-based problems despite having near-perfect scores on others. Second, the main failure mode for count-based problems is not poor structure or calculation errors, but \textit{missing records and hallucinations}. When processing a given message history, the agent occasionally skips transactions, counts the same spending twice under different labels, or creates completely spurious transactions, as depicted in Figure~\ref{fig:4.4_accounting_examples}. This does not mean that the agent hallucinates more often than usual: it processes the vast majority of messages correctly. However, over sufficiently long message sequences, it is likely that at least one skip or hallucination takes place.

Accounting is not the only subset where the agent hallucinates (e.g. see Figure~\ref{fig:4.4_tree_based_examples}, right), but it is the most sensitive. This is because missing (or imagining) a single transaction usually results in an incorrect final settlement. Additionally, the accounting subset contains a significant fraction of unrelated messages, which gives the LLM additional opportunities to hallucinate.
There are several other failure modes and ``quirks'' not specific to memory management: for instance, \texttt{gemini-3.1-pro} and \texttt{gpt-5.5} tend to give more elaborate responses while \texttt{deepseek-v4-pro/flash} often answers in single numbers / sentences. See Appendix~\ref{app:error_analysis} for additional examples.

Curiously, the gap between memory agents with and without hints is far larger than the difference between \texttt{Mem0} and mem-agent.
When evaluating other models in Appendix~\ref{app:alternative_models}, we found that this gap persists. Newer model generations are better at following hints, but they still often fail to properly organize their memory without hints.
After analyzing LLM behavior that led to incorrect answers, we found two main failure modes: either (i) the LLM does not organize its memory, especially when not hinted or (ii) the LLM hallucinates spurious memories, which happens rarely in normal use, but more often when the LLM performs hundreds of consecutive memory updates. See Appendix~\ref{app:error_analysis} for details.


\section{Discussion}\label{sect:discussion}

In this work, we analyzed how modern LLM agents organize their memory. We created StructMemEval, an evaluation benchmark built around problems that require building structured knowledge to solve effectively. Our analysis has shown that retrieval-only systems are unable to solve StructMemEval problems, but more flexible agentic memory systems can achieve significantly higher accuracy, though still far from perfect. We investigated agent trajectories and found that wrong answers are largely caused by suboptimal memory organization and hallucinations.

This also highlights a growing problem in memory-augmented LLMs: as they scale to longer interactions and more conversation turns, previously tolerable hallucination rates become problematic. This suggests two directions for future research: training (or prompting) backbone LLMs to better structure knowledge within existing frameworks, and designing memory systems that facilitate this capability. We hope StructMemEval helps to inform the design of future LLMs and memory frameworks.

\textbf{Limitations:} there are two notable limitations to consider. First, our main evaluations do not use multiple random seeds since encoding multiple hundred conversation turns per sample already accrues significant API costs. Our aggregate results have sufficient sample sizes to draw conclusions, but individual subset-level accuracies can be noisy. Second, our analysis includes multiple proprietary LLMs to evaluate their cutting-edge capabilities. However, proprietary LLMs are known to have a limited availability window and will eventually be deprecated. To alleviate this, we also include \texttt{deepseek-v4-pro/flash} open-weights models that do not rely on proprietary APIs.

\textbf{Acknowledgements.} Authors would like to thank Gleb Rodionov for his help with the evaluation setup and paper presentation. We also thank Olga Vysotskaya for her guidance on accounting practices and terminology. Finally, we thank Denis Mazur, Alexander Rostov, Artyom Eliseev, and Denis Kuznedelev for helpful discussions and brainstorming ideas for benchmark scenarios.

\bibliography{bibliography}
\bibliographystyle{iclr2027_conference}

\appendix
\newpage
\section{Details on Data Collection}\label{app:details_data_collection}

In this section, we describe the data gathering protocol for each task subset in StructMemEval. Our approach consists of the following stages: i) manually create initial scenarios to test their feasibility for different memory types, ii) use LLM augmentations to create multiple similar scenarios and additional evaluation questions, iii) manually verify the resulting scenarios to fix hallucinations and make sure that the reference answers are valid. However, there are slight variations in methodology depending on the task at hand. We also use 3 different LLMs for augmentation to reduce the risk of style contamination implicitly favoring one model over another~\citep{panickssery2024llmevaluatorsrecognizefavor}.

\textbf{Accounting problems:} we start by manually specifying examples of transactions (e.g. ''Alice gave Bob \$N for ...''), then use \texttt{deepseek/deepseek-v3.1-terminus} to generate more transactions similar to the manually created ones. We also introduce unrelated messages, creating scenarios with around 10\%, 30\% and 50\% unrelated messages (5 unique scenarios each, for a total of 15). For these tasks, the only task for the model is to determine the final settlement between parties (with netting) and test if it matches the correct one. We ask for the total settlement amounts multiple times after certain numbers of transactions (conversation depth) to create tasks of different complexity. To determine the reference settlement, we run the same LLM in agentic setup, giving it tools to compute settlement from individual transactions, then verify and correct answers manually. This is because, without tools and verification, even full-context reference LLM sometimes hallucinates transactions.

\textbf{Recommendation problems:} for this subset, we consider tasks where the LLM assistant interacts with a user who watches movies, reads books or attends other activities over a prolonged period of time, after which the agent must decide what activities the user tried more and which ones they liked more on average, expecting an exact numeric answer. This task is difficult for retrieval-augmented LLMs since they can't calculate statistics for top-k search results if the sample size is larger than $k$. To gather these tasks, we follow a similar pattern to the accounting problems, except that we use \texttt{qwen/qwen3-max} to generate dialogues, then verify with \texttt{deepseek/deepseek-v3.1-terminus}.

\textbf{Tree-based problems:} for relations and family tree tasks, we first generate a reference graph corresponding to a family tree where each node corresponds to a person. For simplicity, we ensure that all persons have unique names to avoid confusion. For each graph, we pick pairs of nodes for which there is a unique shortest path between them that has 10 hops (11 nodes visited). Then, we create subsets of graphs containing 10, 20, 30, 40, 50, 100, 150, 200 and 250 links (encoded as messages). We ensure that each subset includes the shortest path between the pair of nodes that we picked previously. After the agent processes these messages, we ask it to find the shortest path. For augmentation, we use the same DeepSeek model as in the accounting subset, but we allow tool use.

\textbf{State tracking problems:} finally, we gather tasks that require the agent to track state changes. Overall, we adopt a similar pipeline to our accounting tasks, but using \texttt{anthropic/claude-opus-4.5} (through Claude Code) for data augmentation. For this evaluation, we start with control tasks with 0 changes (``static'') and consider progressively more complex transitions up to 5 state changes. Tasks with 1 transition have one state change that affects the final outcome, such as with the ``I owe \$X to my neighbor; I moved to a different city Y, do I owe something to my neighbor?'', and subsequent difficulty levels feature multiple such transitions, all of which are necessary to solve the problem. Unlike accounting tasks, we do not reuse conversations between difficulty levels and instead create different unique conversations for each number of transitions. Table~\ref{tab:state-tracking-data-stats} summarizes data collection statistics. Curiously, note that the conversations with 2 state transitions are shorter than 1-transition tasks in terms of the average number of messages, but models still find them more difficult.

\begin{table}[ht]
\centering
\caption{Data gathering statistics for state tracking problems.}
\label{tab:state-tracking-data-stats}
\begin{tabular}{lrrrrrr}
\toprule
Difficulty & Files & Sessions & Total Msgs & User Msgs & Avg. msgs/session & Avg. msgs/file \\
\midrule
static & 7 & 15 & 150 & 75  & 10.0 & 10.7 \\
1tr    & 7 & 35 & 219 & 110 & 6.3  & 15.7 \\
2tr    & 7 & 36 & 188 & 94  & 5.2  & 13.4 \\
3tr    & 7 & 49 & 238 & 119 & 4.9  & 17.0 \\
4tr    & 7 & 59 & 296 & 148 & 5.0  & 21.1 \\
5tr    & 7 & 73 & 362 & 181 & 5.0  & 25.9 \\
\midrule
\textbf{TOTAL} & \textbf{42} & \textbf{267} & \textbf{1453} & \textbf{727} & \textbf{5.4} & \textbf{17.3} \\
\bottomrule
\end{tabular}
\end{table}

\section{Details on Memory Frameworks}\label{app:details_memory_frameworks}

In Section~\ref{sect:experiments}, we evaluate the following LLM configurations:\begin{itemize}[leftmargin=*]
    \vspace{-5px} \item \textbf{Retrieval-augmented LLM:} we considered several implementations for purely retrieval-augmented LLMs and eventually found that \texttt{Mem0}'s built-in retrieval module over OpenAI's \texttt{text-embedding-3-large} embeddings works well on standard benchmarks. To evaluate this configuration, we use \texttt{Mem0} codebase\footnote{\url{https://github.com/mem0ai/mem0}} version 1.0.2, run the built-in search module, then feed the search results into the main LLM to compose the final answer. We search as follows:
\end{itemize}
{\hfill
\begin{minipage}[t]{0.975\linewidth}
\begin{lstlisting}[language=Python]
import mem0
m = mem0.memory.main.Memory(
  mem0.configs.base.MemoryConfig(
    llm=mem0.llms.configs.LlmConfig(provider=..., model=..., api_key=...),  # main LLM
    embedder=mem0.embeddings.configs.EmbedderConfig(
        provider="openai", config={"model": "text-embedding-3-large", "api_key":...}),
    vector_db=mem0.vector_stores.configs.VectorStoreConfig(
        provider="qdrant", config={"collection_name": ..., "path": ...,  # random uid
        "embedding_model_dims": 3072},
    )
  )
)
m.reset()
m.add(FEED_CONVERSATION_HERE, user_id="same_user_everywhere", infer=False)
results = m.search(question, user_id="same_user_everywhere", limit=TOP_K_LIMIT_HERE)
\end{lstlisting}
\end{minipage}\hspace{2px}
}

\begin{itemize}[leftmargin=*]
    \item \textbf{Mem-agent memory:} for a simple abstract memory framework, we use the markdown-based memory from~\cite{mem-agent} using their official codebase\footnote{\url{https://github.com/firstbatchxyz/mem-agent}}. This memory framework allows the agent to create, modify and retrieve any number of markdown files locally. While the original study focuses on teaching small LLMs (\texttt{Qwen3-4B}) to use this memory, we found that \texttt{google/gemini-2.5-pro} can already use the memory well with the prompt alone: it solves simple retrieval tasks well and, when it fails on our tasks, it is because of a knowledge organization error or hallucination, not failure to use the framework itself. Our main rationale for experimenting with Mem-agent is that it is extremely easy to use: pure-python using files, no need to spin up services before use. However, as our experiments demonstrate, it can implement proper memory organization patterns when hinted to do so, which means that it is flexible enough for our tasks.
    When evaluating mem-agent, we instantiate \texttt{mem\_agent = agent.agent.Agent(model=MAIN\_LLM\_HERE)} and use \texttt{mem\_agent.chat(input\_shard)} to both encode memory and answer questions. Crucially, we clear agent's recent message history (\texttt{mem\_agent.messages = mem\_agent.messages[:1]}) after each conversation message. That way, we can be certain that the agent answers the question from memory, not by directly accessing conversation history. This corresponds to a use case where the messages are spread over multiple conversations over a long time-frame and are not neatly packed into one session. \textbf{When evaluating with hints}, we additionally modify the agent's system prompt (first message) and append the task-specific hint after the default mem-agent prompt.
    \item \textbf{\texttt{Mem0} agentic memory:} Finally, we evaluate the  full agentic \texttt{Mem0} memory. For this evaluation, we use the same search module as in our earlier retrieval-based evaluation, but we allow the LLM to decide which memories to store and how to formulate questions for retrieval, allowing the LLM to create, manage and retrieve memories. We use the official codebase~\citep{mem0} with default parameters. \underline{Note that both \texttt{Mem0} and other frameworks will likely work better after tuning.}
\end{itemize}

\textbf{On the (lack of) hyperparameter tuning.} In this initial evaluation, we deliberately evaluate memory as is with minimal tuning. We urge the readers not to draw conclusions such as ``Memory A is better than memory B because of this experiment.'' There are two main reasons for this:\begin{itemize}[leftmargin=*]
\item \textbf{Framework configuration:} Popular long-term memory frameworks typically have a large number of ``knobs'' to tune to optimize the framework for the given task. We avoid this in our initial analysis.
\vspace{-5px}\item \textbf{Model dependency:} we run our experiments with \texttt{gemini-2.5-pro} with some additional experiments with \texttt{gemini-3-pro} and \texttt{gpt-4o-mini} in Appendix~\ref{app:alternative_models}. Other models can score differently.
\end{itemize}

\section{Additional Evaluations for Section~\ref{sect:experiments_lengths}}\label{app:alternative_models}
\subsection{Count-based Tasks}

\begin{table}[ht]
\centering
\caption{Early evaluation for \texttt{gemini-2.5-pro} and different retrieval budgets on count-based tasks for 10\% and 30\% noise (we omit 50\% as non-informative). Agentic memory fares poorly across all budgets due to hallucinations. Gemini 3 pro scores significantly higher (see Table~\ref{tab:app_detailed_accounting_gemini3pro} below).}
\label{tab:app_detailed_accounting_gemini2.5pro}
\begin{tabular}{l|ccc|ccc|ccc}
\toprule
Noise level
& \multicolumn{3}{c|}{10\%} 
& \multicolumn{3}{c|}{30\%} 
& \multicolumn{3}{c}{Average} \\
Transactions & 10 & 20 & 50 
& 10 & 20 & 50 
& 10 & 20 & 50 \\
\midrule
Retrieval (top-5) 
& 0 & 0 & 0
& 0 & 0 & 0 
& 0 & 0 & 0 \\

Retrieval (top-10) 
& 0.6 & 0 & 0
& 0.6 & 0 & 0 
& 0.6 & 0 & 0 \\

Retrieval (top-20) 
& 1.0 & 0.4 & 0
& 0.6 & 0.6 & 0 
& 0.8 & 0.5 & 0 \\

mem-agent (w/o hint)
& 0 & 0 & 0
& 0 & 0 & 0 
& 0 & 0 & 0 \\

mem-agent (with hint)
& 0.2 & 0 & 0
& 0.2 & 0.2 & 0 
& 0.2 & 0.1 & 0 \\
\bottomrule
\end{tabular}

\end{table}

\begin{table}[ht]
\vspace{5px}
\centering
\setlength{\tabcolsep}{5px}
\caption{Detailed evaluation for \texttt{gemini-3-pro} and different retrieval budgets on count-based tasks.}
\label{tab:app_detailed_accounting_gemini3pro}
\begin{tabular}{l|ccc|ccc|ccc|ccc}
\toprule
Noise level
& \multicolumn{3}{c|}{10\%} 
& \multicolumn{3}{c|}{30\%} 
& \multicolumn{3}{c|}{50\%} 
& \multicolumn{3}{c}{Average} \\
Transactions & 10 & 20 & 50 
& 10 & 20 & 50 
& 10 & 20 & 50 
& 10 & 20 & 50 \\
\midrule
Retrieval (top-5) 
& 0 & 0 & 0
& 0 & 0 & 0 
& 0.6 & 0 & 0 
& 0.2 & 0 & 0 \\

Retrieval (top-10) 
& 1 & 0 & 0 
& 1 & 0 & 0 
& 1 & 0.6 & 0 
& 1 & 0.2 & 0 \\

Retrieval (top-20) 
& 1 & 0.6 & 0 
& 1 & 0.6 & 0 
& 1 & 1 & 0 
& 1 & 0.73 & 0 \\

mem-agent (w/o hint)
& 0.8 & 0.6 & 0.2 
& 0.8 & 0.6 & 0 
& 0.6 & 0.8 & 0.4 
& 0.73 & 0.67 & 0.20 \\

mem-agent (with hint)
& 0.8 & 0.8 & 0.4 
& 1 & 0.6 & 0.2 
& 0.8 & 0.4 & 0.4 
& 0.67 & 0.47 & 0.27 \\
\bottomrule
\end{tabular}
\vspace{5px}
\end{table}

\subsection{State Tracking Tasks}

\begin{table}[ht]
\centering
\small

\caption{Accuracy on state tracking tasks by model, \textbf{tr} is the number of relevant state transitions. Per-transition results are noisy due to small sample. Avg. is the mean accuracy over all scenarios.}
\label{tab:app_detailed_state_tracking_eval}
\begin{tabular}{llccccccc}
\toprule
Model & Memory System & 0tr. & 1tr. & 2tr. & 3tr. & 4tr. & 5tr. & Avg. \\
\midrule
\multirow{7}{*}{gpt-4o-mini}
& retrieval (top-5)        & 86\% & 43\% & 0\%  & 14\% & 0\%  & 0\%   & 24\% \\
& retrieval (top-20)       & 86\% & 71\% & 43\% & 29\% & 14\% & 0\%   & 40\% \\
& mem-agent (w/o hint)     & 71\% & 43\% & 86\%   & 29\%   & 71\%  & 43\%    & 57\% \\
& mem-agent (with hint)            & 86\% & 71\% & 86\% & 43\% & 57\% & 100\% & 74\% \\
& \texttt{Mem0} agent (w/o hint)    & 71\% & 57\% & 29\% & 14\% & 0\%  & 0\%   & 29\% \\
& \texttt{Mem0} agent (with hint)   & 43\% & 57\% & 71\% & 71\%& 71\% & 43\%  & 60\% \\
\midrule
\multirow{6}{*}{gemini-2.5-pro}
& retrieval (top-5)        & 86\% & 29\% & 0\%  & 14\% & 0\%  & 0\%   & 21\% \\
& retrieval (top-20)       & 71\% & 71\% & 14\% & 0\%  & 0\%  & 0\%   & 26\% \\
& mem-agent (no hint)      & 57\% & 71\% & 100\%& 71\% & 57\% & 29\% & 64\% \\
& mem-agent (with hint)    & 71\% & 86\% & 71\% & 86\% & 86\% & 71\%  & 79\% \\
& \texttt{Mem0} agent (w/o hint)     & 86\% & 57\% & 71\% & 57\% & 43\% & 57\% & 62\%\\
& \texttt{Mem0} agent (with hint)   & 71\% & 57\% & 86\% & 100\% & 86\%& 86\%  & 81\% \\
\midrule
gemini-3-flash
& \texttt{Mem0} agent (hint)    & 71\% & 100\%& 100\%& 100\% & 100\%& 100\% & 95\% \\
\bottomrule
\end{tabular}
\end{table}


\newpage

\section{Memory Organization Hints from Section~\ref{sect:experiments_hints}}\label{app:memory_organization_hints_prompts}

Below we provide the full memory organization hint for tree-based problems used in Section~\ref{sect:experiments_hints}. The remaining 3 organization prompts for different subsets (State tracking, count-based, recommender systems) can be found in the supplementary code.

\begin{tcolorbox}[
        width=\textwidth,
        colback=white,
        colframe=black,
        boxrule=0.5pt,
        arc=0pt,
        left=6pt,
        right=6pt,
        top=6pt,
        bottom=6pt,
        breakable
    ]

    \textbf{\large Memory Agent System Prompt}

    You are an LLM agent with a self-managed memory system. You interact with memory using Python code blocks.

    \textbf{Response Format}

    EVERY response MUST follow this exact structure:

    \begin{enumerate}[leftmargin=*]
        \item \texttt{<think>} -- Your reasoning (ALWAYS required)
        \item \texttt{<python>} -- Python code OR empty \texttt{<python></python>} (ALWAYS required)
        \item \texttt{<reply>} -- Your response to the user (ONLY when \texttt{<python>} is empty)
    \end{enumerate}

    \textbf{CRITICAL: Always close ALL tags. NEVER put text outside tags. NEVER skip \texttt{<think>}.}

    \textbf{Pattern A: Execute code (wait for results)}

    \begin{lstlisting}[style=prompt]
<think>I need to read memory.</think>
<python>
content = read_file("entities/johnny_fisher.md")
</python>
    \end{lstlisting}

    \textbf{Pattern B: Reply (no code)}

    \begin{lstlisting}[style=prompt]
<think>I found the connection. I can answer now.</think>
<python></python>
<reply>Yes, they are indirect colleagues according to their graph relations.</reply>
    \end{lstlisting}

    \textbf{After Receiving \texttt{<result>} Blocks}

    \begin{enumerate}[leftmargin=*]
        \item Start a new \texttt{<think>} to analyze results
        \item Pattern A for more code, or Pattern B to answer
    \end{enumerate}

    \textbf{Results format:} \texttt{\{'variable\_name': value\}} -- your assigned variables and their values.
    \texttt{\{\}} means you forgot to assign results to variables!

    \hr

    \textbf{\large CRITICAL: Variables Do NOT Persist Between \texttt{<python>} Blocks}

    Each \texttt{<python>} block runs in a FRESH environment. Variables from one block are NOT available in the next.

    \textbf{All read-modify-write MUST happen in ONE \texttt{<python>} block.}

    \hr

    \textbf{\large Memory API}

    \textbf{ALWAYS assign results to variables!}

    \begin{lstlisting}[style=prompt]
content = read_file("entities/johnny_fisher.md")   # CORRECT
read_file("entities/johnny_fisher.md")             # WRONG -- result lost!
    \end{lstlisting}

    \begin{lstlisting}[style=prompt]
create_file(file_path: str, content: str = "") -> bool   # Creates or OVERWRITES
read_file(file_path: str) -> str                          # Returns error string if file missing
check_if_file_exists(file_path: str) -> bool
list_files() -> str                                       # Directory tree
go_to_link(link_string: str) -> str                       # Read [[entities/name.md]] link
    \end{lstlisting}

    \hr

    \textbf{\large Memory Structure}

    \textbf{Entity Files}

    Each person gets ONE file in \texttt{entities/}. File MUST start with \texttt{\# Full Name}.

    \textbf{File naming:} full name in snake\_case + \texttt{.md}
    \begin{itemize}[leftmargin=*]
        \item "Johnny Fisher" -> \texttt{entities/johnny\_fisher.md}
        \item "Kathleen Herrera" -> \texttt{entities/kathleen\_herrera.md}
    \end{itemize}

    \textbf{Entity File Format}

    \begin{lstlisting}[style=prompt]
# Johnny Fisher
## Colleagues
- Christopher Peterson -> [[entities/christopher_peterson.md]]
- Jane Doe -> [[entities/jane_doe.md]]
    \end{lstlisting}

    \textbf{Rules}

    \begin{enumerate}[leftmargin=*]
        \item One \texttt{\# Full Name} header on line 1
        \item One \texttt{\#\# Colleagues} section listing all direct colleagues
        \item Each entry: \texttt{- Full Name -> [[entities/snake\_case\_name.md]]}
        \item \textbf{Symmetric:} if A is in B's colleagues, B MUST also be in A's colleagues
    \end{enumerate}

    \hr

    \textbf{\large Loading Phase: Saving Relationships}

    When the user states "A works with B" or "A is a colleague of B":

    \textbf{Write BOTH entity files in ONE \texttt{<python>} block -- variables don't persist between blocks.}

    \begin{lstlisting}[style=prompt]
person_a = "Johnny Fisher"
person_b = "Christopher Peterson"
file_a = "entities/johnny_fisher.md"
file_b = "entities/christopher_peterson.md"

# Read existing content or initialize
exists_a = check_if_file_exists(file_a)
content_a = read_file(file_a) if exists_a else f"# {person_a}\n## Colleagues\n"

exists_b = check_if_file_exists(file_b)
content_b = read_file(file_b) if exists_b else f"# {person_b}\n## Colleagues\n"

# Add B -> A's file (skip if already present)
if file_b not in content_a:
    content_a = content_a.rstrip('\n') + f"\n- {person_b} -> [[{file_b}]]\n"

# Add A -> B's file (skip if already present)
if file_a not in content_b:
    content_b = content_b.rstrip('\n') + f"\n- {person_a} -> [[{file_a}]]\n"

result_a = create_file(file_a, content_a)
result_b = create_file(file_b, content_b)
    \end{lstlisting}

    \begin{itemize}[leftmargin=*]
        \item Read + modify + write both files in ONE block
        \item Don't split the read of \texttt{content\_a} into one block and the write into another -- \texttt{content\_a} will be gone
    \end{itemize}

    \hr

    \textbf{\large Query Phase: Answering Connectivity Questions}

    When asked "Are X and Y connected/related through colleague relations?":

    \textbf{Follow colleague links one hop at a time.} Read person X's file -> check if Y is in their colleagues -> if not, follow one link to the next person -> repeat until Y is found or no unvisited links remain.

    \textbf{Turn 1 -- Read the starting person's file}

    \begin{lstlisting}[style=prompt]
<think>
Goal: find connection between Johnny Fisher and Brenda Nguyen.
Starting from: Johnny Fisher.
Visited so far: [Johnny Fisher].
</think>
<python>
content = read_file("entities/johnny_fisher.md")
</python>
    \end{lstlisting}

    \textbf{After each result -- Check and hop}

    \begin{lstlisting}[style=prompt]
<think>
Johnny Fisher's colleagues: Christopher Peterson.
Target (Brenda Nguyen) not found yet.
Next: Christopher Peterson (unvisited).
Visited so far: [Johnny Fisher, Christopher Peterson].
</think>
<python>
content = go_to_link("[[entities/christopher_peterson.md]]")
</python>
    \end{lstlisting}

    \textbf{When the target appears -- Answer immediately}

    \begin{lstlisting}[style=prompt]
<think>
Joan Morgan's colleagues: Brenda Nguyen -> [[entities/brenda_nguyen.md]].
Target found. Path confirmed: Johnny Fisher -> ... -> Joan Morgan -> Brenda Nguyen.
</think>
<python></python>
<reply>Yes, they are indirect colleagues according to their graph relations.</reply>
    \end{lstlisting}

    \textbf{State to track in \texttt{<think>} at every hop}

    Always write these three lines in \texttt{<think>} before each hop:
    \begin{itemize}[leftmargin=*]
        \item \textbf{Goal:} the person you're looking for
        \item \textbf{Visited:} full list of nodes you've already read (never follow a visited link)
        \item \textbf{Next:} the unvisited colleague you will read next
    \end{itemize}

    This prevents infinite loops and keeps the traversal correct.

    \hr

    \textbf{\large Correct vs Wrong -- Traversal}

    \begin{itemize}[leftmargin=*]
        \item Read Johnny Fisher's file, note "Brenda Nguyen not in colleagues", hop to Christopher Peterson
        \item Track visited nodes in \texttt{<think>} every step
        \item Answer as soon as the target name appears in any colleague list
        \item Answer without reading memory
        \item Follow a link to a person already in the Visited list
        \item Give up after one hop if the target isn't a direct colleague
    \end{itemize}

    \hr

    \textbf{\large Operating Rules}

    \begin{enumerate}[leftmargin=*]
        \item \textbf{ALL read-modify-write in ONE \texttt{<python>} block} -- variables don't persist
        \item \textbf{Always assign results} -- unassigned calls return \texttt{\{\}}
        \item \textbf{Symmetric storage} -- always write BOTH entity files per relationship
        \item \textbf{Track visited in \texttt{<think>}} -- list every node you've read, never re-read one
        \item \textbf{Valid Python only} -- syntactically correct code
        \item \textbf{Wait for \texttt{<result>}} -- after submitting code, wait before continuing
        \item \textbf{Exact names} -- use exact full names from user messages; construct file paths as snake\_case
    \end{enumerate}

    \hr

    \textbf{\large Filtering}

    In the user query, you might receive a question that includes \texttt{<filter>} tags. You MUST follow these filters exactly. If no filters are present, return the answer as is.

    \end{tcolorbox}

\section{Additional Details for Error Analysis}\label{app:error_analysis}

On top of the specific memory organization problems highlighted in Section~\ref{sect:experiments_error_analysis}, we identified three categories of errors in the agentic memory files for the most demanding high precision benchmark subset -- accounting problems: 1) the omission of transactions, which was the most common; 2) the duplication of existing transactions; and 3) most significantly, the hallucination of transactions, as illustrated in Table~\ref{tab:app_error_analysis}.

\begin{table}[ht]
\centering
\caption{Example memory entries for mem-agent generated for accounting problem 11 with \texttt{gemini-2.5-pro}. There is a transaction with the note ``sandwich'' -- but no line in this conversation contains the word ``sandwich'', and there is no real transaction with its listed payer/amount. Moreover, the first and last  rows are duplicate entries for the same transaction.}
\label{tab:app_error_analysis}
\vspace{5px}
\small
\begin{tabular}{llccccccc}
\toprule
Payer & For Whom & Amount & Currency & Notes \\
\midrule
\\
\textit{Bob}   & \textit{All}      & \textit{57}     & \textit{EUR}      & \textit{lunch} \\
Bob   & All      & 51     & EUR      & snacks \\
Alice & Alice, Bob & 84 & EUR & dinner \\
Alice & All & 120 & EUR & train tickets \\
Charlie & Bob, Charlie & 60 & EUR & groceries \\
Alice & All & 45 & EUR & Museum entry \\
Charlie & All & 75 & EUR & Lunch \\
Alice & Alice, Charlie & -15 & EUR & refund on museum tickets \\
Bob & All & 30 & EUR & taxi \\
Alice & All & 90 & EUR & Dinner \\
Bob & Alice, Bob & 36 & EUR & breakfast \\
\textbf{Alice} & \textbf{Bob} & \textbf{15} & \textbf{EUR} & \textbf{sandwich} \\
\textit{Bob}   & \textit{All}      & \textit{57}     & \textit{EUR}      & \textit{lunch} \\
\end{tabular}
\end{table}

Additionally, we found that different LLM backbones have idiosyncrasies about how they choose to formulate their responses. \texttt{Deepseek-v4-flash} prefers to give short answers where possible, typically replying with individual numbers or short phrases if the question allows it. In turn, \texttt{gemini-3.1-pro} and \texttt{gpt-5.5} are significantly more likely to answer with one or multiple paragraphs. An extreme case of this trait is \texttt{gpt-5.5-pro} that often leverages extensive real-world knowledge about locations, movies, and other related matters --- to an extent where it sometimes prefers that knowledge over personalized user-related memories.

\begin{figure}[h]
    \centering
        \begin{tcolorbox}[
    examplebox, 
    title={Verbose \texttt{gpt-5.5-pro} response that leverages real-world knowledge instead of memory.},
    ]
\textbf{Query}
\vspace{2pt}

\begin{quote}
\ttfamily
What should I do this Saturday based on where I live now?
\end{quote}

\vspace{4pt}

\textbf{LLM Response}
\vspace{2pt}

\begin{quote}
\footnotesize
Since you live in Capitol Hill now, I'd keep it local and avoid the tourist-heavy stuff:

\begin{itemize}[leftmargin=1.2em]
    \item \textbf{Morning:} Grab coffee at \textbf{Espresso Vivace} or \textbf{Victrola}, then walk to \textbf{Volunteer Park}.
    \item \textbf{Late morning:} Visit the \textbf{Seattle Asian Art Museum}.
    \item \textbf{Lunch:} Eat at \textbf{Plum Bistro} or \textbf{Aviv Hummus Bar}.
    \item \textbf{Afternoon:} Browse \textbf{Elliott Bay Book Company}.
    \item \textbf{Evening:} Check live music at \textbf{Neumos}, \textbf{Barboza}, or \textbf{Chop Suey}.
\end{itemize}

If it's rainy: coffee $\rightarrow$ bookstore $\rightarrow$ museum $\rightarrow$ dinner $\rightarrow$ live show.

If it's sunny: Capitol Hill $\rightarrow$ Volunteer Park $\rightarrow$ Interlaken Park.
\end{quote}

\vspace{4pt}

\textbf{Ground Truth / Expected Behavior}
\vspace{2pt}

\begin{quote}
\footnotesize
The user currently lives in \textbf{Alfama, Lisbon}. A correct response should recommend browsing the \textbf{Feira da Ladra} flea market for vintage records.

The answer should \emph{not} reference activities from previous cities such as:
\begin{itemize}[leftmargin=1.2em]
    \item Kungsholmen kayaking on Riddarfjärden
    \item Sea Point morning runs
    \item Shoreditch and Columbia Road Flower Market
    \item Williamsburg and Smorgasburg
    \item Montmartre sketching at Place du Tertre
\end{itemize}
\end{quote}
    \end{tcolorbox}
\end{figure}

\section{Broader Impacts}\label{app:broader_impacts}

Our primary contributions are a dataset and an analysis of LLM agents on that dataset. Thus, we believe that there is little direct societal impact from the use of our artefacts and findings. That said, there may be indirect ramifications if our benchmark contributes to future memory agent designs.
Specifically, if our work indirectly helps create memory agents that are better capable of organizing their knowledge, this could impact several use cases, both positive or negative.

\textbf{Positive.} Our evaluation scenarios are based on real personal assistant and agentic use cases: navigating corporate or institutional structures, keeping track of personal expenses, finding useful patterns from managing a codebase or a server over extended amounts of time, and others. These are all valuable usage scenario where a ``better-organized'' LLM agent could add more value to its users.

\textbf{Negative.} The ability to organize knowledge could potentially help certain harmful scenarios: tracking or ``doxing'' persons against their will, finding and systematically exploiting vulnerabilities in networks, and potentially other misuse / abuse scenarios that involve drawing conclusions from long streams of events.




\end{document}